# A Low Computational Approach for Price Tag Recognition

M.A. Aliev[1,2], D.A. Bocharov[3], I.A. Kunina[1,3], and D.P. Nikolaev[1,3]
[1]Smart Engines Service LLC, Moscow, Russia
[2]Federal Research Center "Computer Science and Control" RAS, Moscow, Russia
[3]Institute for information transmission problems (Kharkevich Institute) RAS, Moscow, Russia

## ABSTRACT

In this work we discuss the task of search, localization and recognition of price zone within a photograph of the price tag. The task is being addressed for the case when image is acquired by small-scale digital camera and calculation device has significant resource constraints. The proposed approach is based on Niblack binarization algorithm, analysis and clasterization of connected components in conditions of known price tag geometrical model. The algorithm was tested on a private dataset and has shown high quality.

**Keywords**: price tag recognition, local adaptive binarization, connected components search.

## 1. INTRODUCTION

Automatic price tag recognition is in high demand due to the increasing automation of all processes in different business solutions.

The increased level of automation not only reduces the number of human errors but also leads to personnel optimization, use of less complex tools for daily tasks, and reduction of acquisition and running costs for portable data terminal (PDT).

For instance, some retailers already employ price tag recognition for automation of routine inventory tasks [1], using robots for collecting information and control of their products [2], [3]. Besides products control can be performed by the retailer's staff using their own smartphones [4].

On the other hand, recognition of price tags can improve customer satisfaction through price control, or can be used for monitoring competitors' prices and making pricing decisions.

Apart from recognition quality, also in high demand is system's operation in conditions of minimal energy usage. This is due to the fact that utilization of energy and resource demanding algorithms [5] [6] will lead to quick discharging of exploited equipment (robots or devices used by personnel), that will negatively affect the efficiency of the whole system. At the same time this minimal energy usage requirement does not allow the usage of client-server architecture for this task.

The purpose of this work is searching, localization and recognition of retail price on a price tag image. The practically significant problem statement will be considered, when the computing device such as Odroid XU4 or similar is used for price tag recognition with significant constraints on shooting conditions. The paper does not propose some general universal solution, but only the solution to the specific problem with specific constraints (see 2.1). The focus is specifically on recognition of the price and not all the elements of the price tag because the following workflow is used quite often: the placement of goods and their price tags is known and we only need to recognize the price from the price tag to be able to compare it to an external price list (or other data source). Then this comparison can be used by the retailer as a basis for price change or correction if needed.

This work is structured as follows: in section 2 aspects of the task are considered and different existing approaches for its solution reviewed. In section 3 the stages of the proposed algorithm are described. In section 4 the testing results are presented and discussed, and in sections 5 and 6 suggestions for further research and conclusion are considered.

## 2. ASPECTS AND RELATED WORK

### 2.1 Aspects of the task

Let us consider the features of the task:

1. The price tag occupies most of the frame.
2. The price tag is subject to specific filling rules which are known in advance and vary insignificantly.
3. All input image sizes deviate slightly from certain limit.
4. The price tag image comes from smartphone camera or robotic platform frontal camera.
5. Rejection of a correct image by the system is better than accepting an incorrect answer. Wrong answer causes potentiallosses: printing new price tags, wrong price list formation, excess personnel work.

Let us now consider the fourth point of the list and aspects of acquired image.

The use of moving platforms to collect price tags images carries with it all the problems of optical document recognition where images were made using small-scale digital cameras [7]. The following can be noted as possible problems (sometimes combined) in price tag recognition task:

1. Flares
2. Blurred images
3. Noises
4. Price tag surroundings: different deal announcements and complex background because of the goods behind it
5. Tilted or skewed image

In addition to all the mentioned problems the photograph can just be made incorrectly and there will be no price tag on the image or only its fragment will be present.

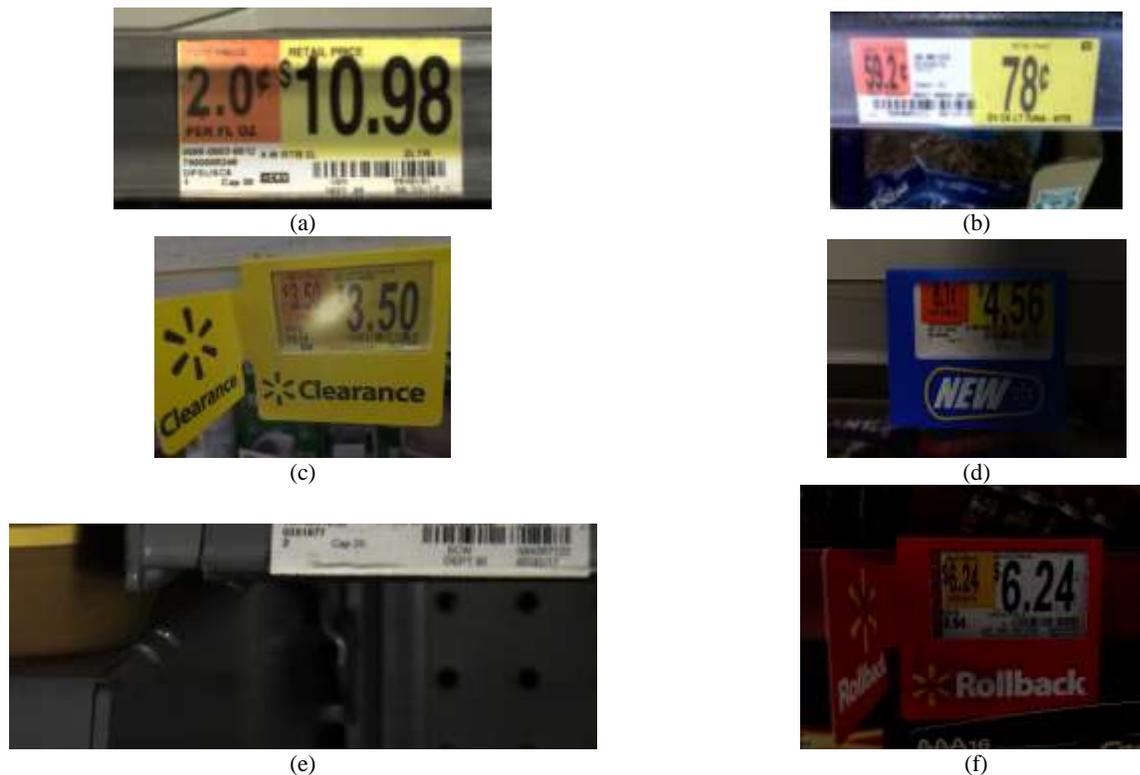

Figure 1. Examples of different price tag recognition problems: (a) low contrast; (b) complicated background and blur; (c) low contrast and flare; (d) low contrast and noise; (e) price tag absence; (f) low contrast and noise.

One more problem is the possible presence of projective distortions but in this work it will be considered insignificant, because when smartphones or robotic platforms are used the shooting parameters and rules are strictly defined once in the beginning of work process. Examples are shown on Figure.1.

As an object of research we chose the price tags dataset some samples out of which are presented in Figure.1. The dataset contains 5 types of price tags, all of which are presented on the Figure.1a,b,c,d,f.

**2.2 Related work**

The following scheme is typical for a recognition task: first, localization of the target zone and then its normalization, next the recognition of normalized zone and post-processing of results. If the target zone is not found or recognition result can not be corrected in line with the given format at post-processing stage then the system rejects the image.

In the last several years the leading positions in the area of text detection and recognition are occupied by neural networks methods [8] [9], but they have some drawbacks [5].

Publicly available and marked up data sets are addressed to recognition in the wild, which is, firstly, significantly wider and because of that much more complicated than our considered task, and, secondly, requires much more computational resources [6].

In addition, readily available models are either excessive for the described constrained problem or require adaptation for specific task (for example, using transfer learning [10]). On the other hand, training "from scratch" needs considerable amounts of ideal images and markup (for example [11] or [12]).

Next, there are approaches that use binarization [13]: for example, in natural images in [14]. However, despite the significant advances made in this area it is problematic to figure out the same parameters for all text extraction because of very different shooting conditions and problems caused by that.

As one more type of approaches saliency maps can be used, for example [15]. But this approach does not take into account such problem aspects as described in section 2.1.

For price tags with clear borders approaches of detecting zone of interest like [16] can be used, followed by the text segmentation and OCR, but it will be necessary first to localize text in the detected zone.

Among the works specifically about price tags the approaches using HSV color space in [17] and neural networks in [18] can be mentioned. But the first work (also as [15]) does not take into account such problem aspects as described in section 2.1, and the second one (just as other tasks that use neural networks) needs large amount of markup for network training.

## 3. ALGORITHM

It is supposed that all input images have sizes within certain limit, otherwise input image is scaled to preassigned size.

### 3.1 Control flowchart

Let us distinguish the target elements (elements of retail price) and noise elements (labels, codes, barcodes, artifacts of image processing) that altogether form full price tag filling.

The control flowchart of proposed algorithm is presented at Figure.2. The main stages of retail-price recognition on price tag images (that are typical for all automatic input systems) are: search and localization of target zone on the image (everything before finding angle), preparing founded zone for recognition (finding angle), recognition (OCR), postprocessing of recognition results.

For the recognition we use a lightweight segmentation and recognition neural network [19] [20] without any special training and setup for price tags fonts. Post-processing is called to normalize recognized value to known price tag format [21].

In this work the first two stages will be considered in detail.

### 3.2 Binarization

Uneven lightning conditions can lead to an uneven brightness of the image. To compensate this it was decided to use Niblack local adaptive binarization algorithm [22].

The goal of the binarization at this stage is the correct binarization of most of the retail price digits area. We don't need to acquire absolutely sharp digits contours at this point because more precise location and sizes of retail price digits will be estimated at later stages.

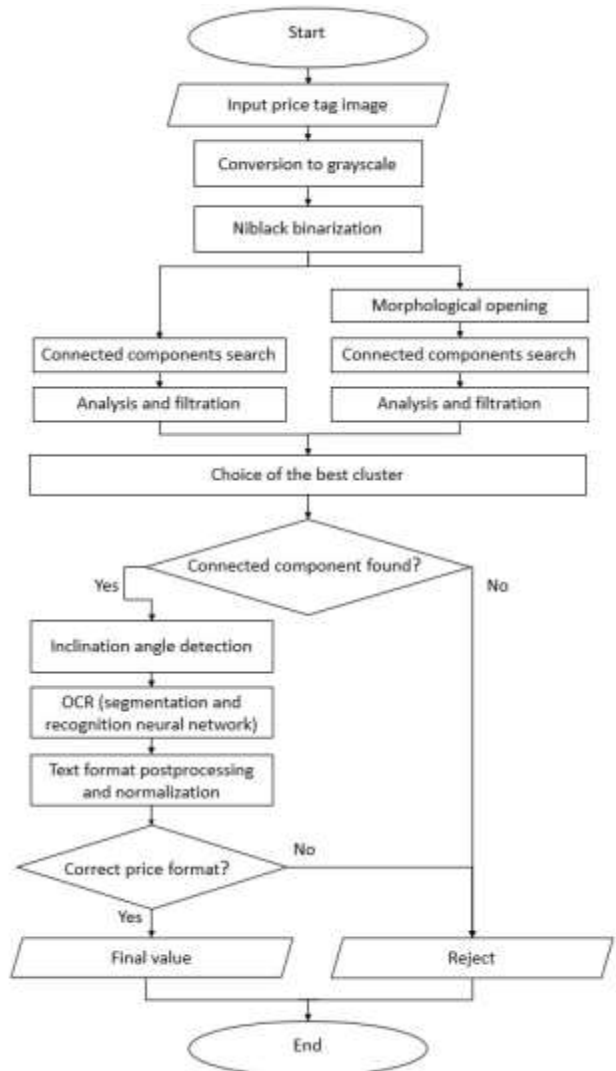
Figure 2. Control flowchart of the proposed algorithm

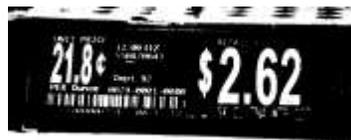
Figure 3. Example of binarized image

Since it is expected that price tag occupies most of the image, it is possible to estimate symbols sizes.

Based on this and described goal of the binarization we can choose binarization parameters such as window sizes. Retail price is located near price tag horizontal borders. That is why vertical window size is chosen as a little more than one digit estimated height. In horizontal surroundings of the retail price there is no comparably sized objects, so horizontal window size is chosen as equal to estimated width of several digits. Example of binarized image is on Figure.3.

### 3.3 Algorithm branching

After binarization we call the morphological operation of opening using square structure element. This is done to suppress possible noise elements on the binarized image while keeping the target elements. The structure element size is chosen accordingly.

However, if the input image is very blurred and has a lot of noise such operation will only make it worse (see Figure.4) - target elements will become even more thin, the gaps will appear or become bigger.

For this reason here the algorithm branches out in two paths, and for further stages two different images come in parallel: the one processed with morphological operation and the original binarized one.

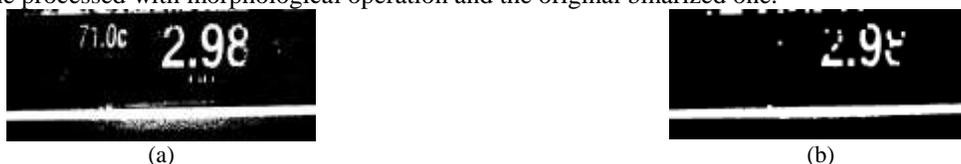

(a) (b)

Figure 4: Example of the image (a) before applying the morphological operation of opening and (b) after. It can be seen that on (b) digits 2 and 8 fall into several small parts.

### 3.4 Connected components analysis

After the branching, connected components with 8-connectivity are calculated on each image. Then their analysis is performed in two steps.

#### 3.4.1 Connected components analysis based on their sizes

At the first step a crude primitive filtration is made based on price digits approximate sizes, their approximate areas and relative ratios of width and height. It is important to have rather soft restrictions not to filter out needed components. Hence after the filtration only such components are left that are candidates for retail price elements (digits).

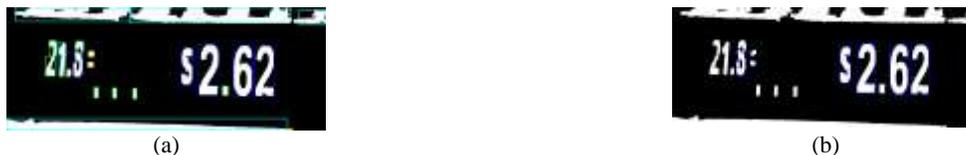

(a) (b)

Figure 5: Example of (a) selecting all found connected components and (b) only those that are left after the first stage filtration.

#### 3.4.2 Connected components analysis based on possible price formats

At the second step the clusterization is performed taking into account possible price formats. After that and also taking into account possible price formats, bounding rectangle for price is calculated.

Knowing the formats allows us to search for clusters with constraints on the number of elements and relative location. It also allows us to verify the correctness of the first stage filtration. For instance knowing that the price can contain a dot (see Figure.1a,c,d,f) it is possible to look for connected component which corresponds to a dot and then use that information further. For example (see 1c) last digit may be not found (because of noise and blur the connected component can fall to pieces), but format knowledge (i.e. that after a dot there are always two digits) allows us to widen the bounding rectangle for one more digit size width.

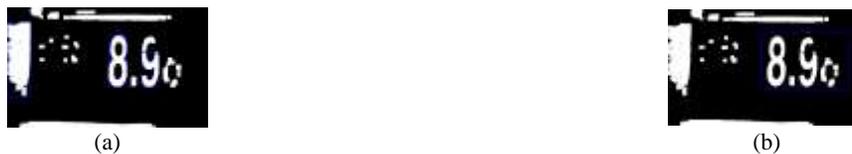

(a) (b)

Figure 6: Example of (a) the candidates to price elements and (b) final bounding rectangle.

### 3.5 Choosing the best cluster

At this stage the comparison of the clusters found in two parallel branches of the algorithm is performed taking into account the price tag format and relative locations of the elements (layout). For example in Figure.1a,b both digit zones on the left and right can be found as the desired region, but the knowledge of the layout and sizes of price tag elements helps to choose the correct one.

### 3.6 Finding inclination angle and its compensation

Due to the capturing conditions the image incline can only be rather small and therefore does not influence the stages of binarization, morphological processing and connected component analysis. But at the recognition stage such inclined

(or even worse, truncated) zone of price will lead to recognition errors. To calculate the image incline angle the fast Hough transform was used as described in [23]. If the angle is larger than the given threshold value the bounding rectangle coordinates are corrected. As a result instead of the bounding rectangle we acquire the bounding quadrangle.

Next, if needed, the inclination angle is compensated while transforming quadrangle into rectangle.

For the next stage it is necessary to have a cropped color image of the retail price region. Because of this, the price zone rectangle cropping or quadrangle transformation to rectangle and its further cropping is performed on color image.

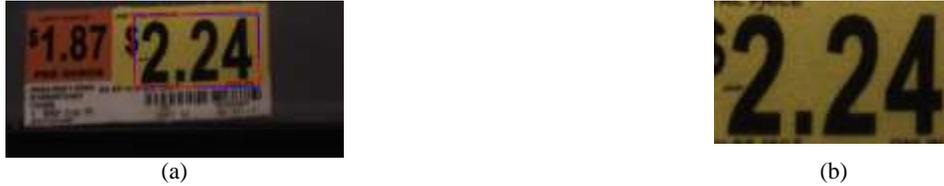

(a) (b)

Figure 7: Examples of (a) bounding rectangle before compensating the incline angle (blue) and after (red). At (b) result of price zone region after inclination angle compensation and cropping.

## 4. TESTING OF THE PROPOSED ALGORITHM

For testing of the proposed approach the dataset from one specific retailer was used; settings for the algorithm steps were made accordingly.

### 4.1 Dataset

The dataset consisted of 708 images, out of which 29 images were incorrect (there were no price tags), and among the remaining correct 679 there were approximately 80 angled images, 90 blurred, 150 with low contrast and 50 washed out.

Image sizes were evenly distributed between $1350 \times 700$ and $800 \times 400$.

Testing was conducted on the full dataset as well as the set of the correct images only.

### 4.2 Results

Let us introduce terminology: True Positive (TP) - number of correct images on which zone was found correctly, True Negative (TN) - number of incorrect images which were correctly rejected, False Positive (FP) - number of correct images on which the target zone was found incorrectly or incorrect images that were accepted, False Negative (FN) - number of correct images which were mistakenly rejected.

Results of the detection for the set of correct images only are shown below in Table 1:

Table 1. Results for the set of correct images only

| Total | TP | | TN | | FP | | FN | | Precision | Recall | Accuracy |
|---|---|---|---|---|---|---|---|---|---|---|---|
| | # | % | # | % | # | % | # | % | | | |
| 679 | 664 | 97.791 | 0 | 0.0 | 13 | 1.915 | 2 | 0.295 | 0.981 | 0.997 | 0.978 |

Results of the detection for the full dataset are shown below in Table 2:

Table 2. Results for the full dataset

| Total | TP | | TN | | FP | | FN | | Precision | Recall | Accuracy |
|---|---|---|---|---|---|---|---|---|---|---|---|
| | # | % | # | % | # | % | # | % | | | |
| 708 | 664 | 93.785 | 15 | 2.119 | 27 | 3.814 | 2 | 0.282 | 0.961 | 0.997 | 0.959 |

Result of retail price recognition for correctly found zones are shown in Table 3. Detection result precision was 96.1%, recall was 99.7%.

Table 3. Result of retail price recognition for correctly found zones.

| Correct zone found | Correct value | Correct value, % |
|---|---|---|
| 664 | 650 | 97.89 |

One of the drawbacks of this algorithm is the increasing number of false positives (FP) when incorrect images are added to the data set. It is also important, however, that in this case the number of true negatives (TN) is higher than the increase in FP.

It is also important to mention that the average time for image recognition (from the input to final price value) is 42 milliseconds on AMD Ryzen$^{TM}$ 5 1600 Processor with base clock 3.2GHz. For comparison, on ICDAR2015 dataset (with resolution 1280x720) in work [24] the end-to-end inference time is 210 ms for one and 330ms for another neural network on a Tesla V100 GPU. In [25] on a Titan-X GPU detection and recognition time for the fastest suggested variant is 59ms and 75ms for the best performing variant.

In case of very difficult and complicated tasks such instruments can be very helpful, but the problem described in this work can be solved with much smaller amount of resources compared to the creation of a new network (research and creation of new network architecture, markup of data, etc).

## 5. FURTHER RESEARCH

In further work the authors plan to expand the test dataset in two directions:
1. Increase the number of incorrect images in order to measure quality of their processing more precisely and improve it, 2. Add more price tags of different formats and types. In that case it will be necessary to research the best way to choose the needed price tag format from the available alternatives. Possible solution here can be the dynamic programming approach suggested in [26].

It will also be interesting to apply less known but probably more suitable for the investigated task binarization methods such as described in [27] and [28].

Another possibility for improving the detector quality is search and recognition of barcodes if they are present, or recognition of the price within a video stream and integration of the results as described in [29].

## 6. CONCLUSION

In this work we proposed a practical solution to the task of search, localization and recognition of price tags within an image. The proposed approach is based on Niblack binarization algorithm, analysis and clusterization of connected components, and post-processing of results. Testing was performed on a dataset consisting of 708 images, with the following results: accuracy is 95.8%, false positives - 3.8%, precision - 96.1%, recall - 99.7%.

## ACKNOWLEDGMENTS

This work is partially financially supported by Russian Foundation for Basic Research (projects 19-29-09092, 18-0701387).